%% file: arxiv.tex
\documentclass{article}

\usepackage[english]{babel}

\usepackage[letterpaper,top=2cm,bottom=2cm,left=3cm,right=3cm,marginparwidth=1.75cm]{geometry}

\usepackage[utf8]{inputenc}
\usepackage[T1]{fontenc}
\usepackage{microtype}

\usepackage{amsmath,amsthm,amsfonts}
\usepackage{graphicx}
\usepackage[colorlinks=true, allcolors=blue]{hyperref}
\usepackage{natbib}
\usepackage{tikz}
\usepackage{pgfplots}

\pgfplotsset{compat=1.10}
\usepgfplotslibrary{fillbetween}
\usetikzlibrary{patterns}

\usepackage{subcaption}
\usepackage{xcolor}

\usepackage{wrapfig}
\usetikzlibrary{calc} 

\definecolor{DarkGreen}{rgb}{0.1,0.5,0.1}
\definecolor{DarkRed}{rgb}{0.5,0.1,0.1}
\definecolor{DarkBlue}{rgb}{0.1,0.1,0.5}
\hypersetup{
	colorlinks=true,       
	linkcolor=DarkBlue,         
	citecolor=DarkGreen, 
	filecolor=DarkBlue,  
	urlcolor=DarkBlue,   
	pdftitle={},
	pdfauthor={},
}

\input{header}

\title{Look-Ahead Reasoning on Learning Platforms}

\usepackage{authblk}
\stepcounter{footnote}
\addtocounter{footnote}{-1}
\author[1]{Haiqing Zhu}
\author[2]{Tijana Zrnic}
\author[3,4]{Celestine Mendler-D\"unner}
\affil[1]{Australian National University}
\affil[2]{Stanford University}
\affil[3]{ELLIS Institute Tübingen}
\affil[4]{Max Planck Institute for Intelligent Systems, T\"ubingen, and T\"ubingen AI Center}
\date{}        
\renewenvironment{abstract}
 {\small
  \begin{center}
  \bfseries \abstractname\vspace{-.5em}\vspace{0pt}
  \end{center}
  \list{}{%
    \setlength{\leftmargin}{20mm}
    \setlength{\rightmargin}{\leftmargin}%
  }%
  \item\relax}
 {\endlist}

\begin{document}

\maketitle

\begin{abstract}
On many learning platforms, the optimization criteria guiding model training reflect the priorities of the designer rather than those of the individuals they affect. Consequently, users may act strategically to obtain more favorable outcomes. While past work has studied strategic user behavior on learning platforms, the focus has largely been on strategic responses to a deployed model, without considering the behavior of other users. In contrast, \emph{look-ahead reasoning} takes into account that user actions are coupled, and---at scale---impact future predictions. Within this framework, we first formalize level-$k$ thinking, a concept from behavioral economics, where users aim to outsmart their peers by looking one step ahead. We show that, while convergence to an equilibrium is accelerated, the equilibrium remains the same, providing no benefit of higher-level reasoning for individuals in the long run. Then, we focus on collective reasoning, where users take coordinated actions by optimizing through their joint impact on the model. By contrasting collective with selfish behavior, we characterize the benefits and limits of coordination; a new notion of alignment between the learner's and the users' utilities emerges as a key concept. 
Look-ahead reasoning can be seen as a generalization of algorithmic collective action; we thus offer the first results characterizing the utility trade-offs of coordination when contesting algorithmic systems.~\looseness=-1
\end{abstract}


\section{Introduction}

Digital platforms deploy learning algorithms that collect and analyze data about individuals to power services, personalize experiences, and allocate resources.
As people come to understand how these systems make decisions, they often adapt strategically to improve their outcomes.

Prior research has largely modeled such strategic behavior as \emph{unilateral}: each agent responds to the platform's decision rule by optimizing their own outcome while treating that rule as fixed. For example, a job applicant might rephrase their resume to include keywords that align with an automated screening system's preferences. This perspective neglects the fact that many others may be doing the same---thereby collectively shifting the data distribution from which the platform learns in the future. 

In reality, there is ample empirical evidence that users frequently reason about one another's behavior. They may act in solidarity~\citep{tassinari20solidarity}, anticipate other people's adaptations to gain an advantage~\citep{kneeland15}, or  coordinate to amplify their collective influence~\citep{chen18didi}, in the latter case oftentimes facilitated by labor organizations \cite[e.g.,][]{drivers-united}. %
On learning platforms in particular, such reasoning involves not only anticipating the behavior of other platform participants but also how those behavioral changes will impact the learning algorithm in the future. We call this \emph{look-ahead reasoning}.  In the resume-screening example, look-ahead reasoning might surface as choosing to emphasize distinct keywords that others have abandoned, anticipating that popular buzzwords will lose predictive value as they become widespread.

\subsection{Our contributions}

We characterize how look-ahead reasoning---user behavior that anticipates the actions of others in the population---reshapes learning dynamics and equilibria on learning platforms. We begin with selfish agents, who act independently but strategically account for the other agents' responses to the deployment of a predictive model. We then turn to coordinated behavior, where agents act jointly and strategically leverage their influence on the learner. Finally, contrasting the two settings allows us to characterize the benefits and limitations of coordination on learning platforms.

To capture agents who selfishly aim to outsmart their peers, we formalize the concept of \emph{level-$k$} thinking~\citep{nagel95} from behavioral economics in the context of learning systems. Level-$k$ thinking captures different depths of strategic thought: a level-$k$ thinker acts assuming they are ``one step ahead'' of all other individuals in the population, who are level-$(k-1)$ thinkers. A level-$0$ thinker is non-strategic. Higher levels $k$ are defined recursively. In our setup, actions are determined by hallucinating the data distribution resulting from the assumed behavior of other agents and best-responding to the predictive model it induces. We study the dynamics of repeatedly retraining a model acting on a population of level-$k$ thinkers. We show that ``deeper'' thinking achieved for larger $k$ accelerates the learning dynamics, while resulting in the \emph{same} equilibrium solution, no matter the depth of thinking.

\begin{theorem}[Informal]
    For $k\geq1$, let $\alpha_k\in (0,1)$ be the fraction of level $k$-thinkers in the population, $\sum_{k=1}^\infty \alpha_k=1$. Assume the learner minimizes a loss function that is smooth and strongly convex, and suppose that the agent responses are sufficiently Lipschitz in the model parameters. Then, for some constant $c\in(0,1)$, repeated retraining converges to a unique stable point at rate \[O\left(\Big[\sum_{k=1}^{\infty} c^{k} \alpha_{k}\Big]^t\right).\] 
\end{theorem}

Therefore, selfish behavior, even if it relies on higher levels of reasoning, does not improve the agents' utility in equilibrium. 

Next, we investigate whether agents can move past this obstacle if they coordinate. 
Look-ahead reasoning with coordination allows steering model updates by strategically and systematically modifying the data that the learner learns from.
We show that the gap between coordination and lack thereof in terms of agent utility is governed by a notion of alignment between the objectives of the learning platform and the population. Below, we use $\ell(z,\theta)$ and $u(z,\theta)$ to denote the learner's loss and the agent utility for deploying model $\theta$ on instance $z$.  Furthermore, $\langle a,b\rangle_M  := a^\top M b$.

\begin{theorem}[Informal]
    Let $\cD^*$ and $\cD^\sharp$ denote the population's data distributions at equilibrium under selfish reasoning and under collective reasoning, respectively. Then, assuming regularity conditions, the benefit of coordination $\mathrm B$, defined as the difference in population utility at the two equilibria, is bounded as
    \[\mathrm{B} \leq \left(\innerH{\expectsub{z\sim \cD^*}{\nabla_\theta u^*}}{\expectsub{z\sim \cD^\sharp}{\nabla_\theta \ell^*}}\right)^2,\]
    where $\mathbf{H}^\star =\expectsub{z\sim {\cD^*}}{\nabla_\theta^2 \ell(z,\theta^*)}$ and $\theta^*$ denotes the equilibrium model under selfish reasoning. We use the short-hand notation $\nabla_\theta u^* = \nabla_\theta u(z, \theta^*)$, $\nabla_\theta \ell^* = \nabla_\theta \ell(z, \theta^*)$.
\end{theorem}

Thus, if the average agent utility and the average loss of the learner are orthogonal, there is no benefit to coordination. Similarly, when $u=c\cdot \ell$ for either $c>0$ or $c<0$, the benefit of coordination is zero. However, when there is the right amount of overlap between the objectives that the collective can exploit, coordination can lead to more favorable outcomes than selfish reasoning. For example, think of instance $z$ as a feature-label pair $(x,y)$ and consider a learner and a collective who both care about predictions $f_\theta(x)$: the learner cares about making them accurate, and the collective cares about them having certain favorable values. Then, modifying the labels $y$ can be an effective strategy for the collective to steer the predictive model $\theta$ through the learner's response, something that individuals reasoning selfishly cannot achieve. This overlap in the two sides' objectives enables a large benefit of coordination. We elaborate on this example later on.

In additional results, we study heterogeneous populations made up of selfish agents and collectives of varying size. The results explain why larger collectives do \emph{not} always lead to a higher utility for participating agents. They also show how broader participation in the collective stabilizes learning dynamics.~\looseness=-1

\subsection{Background and related work}

Strategic classification~\citep{hardt16strategic,bruckner2011stackelberg} introduces a model to study strategic behavior in learning systems based on assumptions of individual rationality. It describes a population of agents best-responding to a decision rule by altering their features to achieve positive predictions, given a fixed decision rule. Several variations of this basic model have been studied \citep[e.g.,][]{dong2018strategic, chen2020learning, bechavod2021information, ghalme2021strategic,jag21alt, levanon22ageneralSC}; see \cite{podimata2025incentive} for a recent survey of this literature. All these works focus on studying how agents strategize against a fixed decision rule. Our work introduces a new dimension of reasoning to strategic classification, taking into account how individual agents' actions are coupled and how this influences the predictive model the agents strategize against.

Performative prediction \citep{perdomo20performativeprediction} introduces performative stability as an equilibrium notion that characterizes long-term outcomes in the interaction of a population with a learning system. Performative stability is a fixed point of repeated retraining by the learner in a dynamic environment. Prior work in performative prediction~\citep[e.g.,][]{perdomo20performativeprediction,mendler20stochasticPP,drusvyatskiy23stochastic,brown22stateful, narang23multiplayer} has studied the behavior of retraining and conditions that ensure its convergence to stability in different learning settings. We refer to \cite{hardt25sts} for a more extensive overview of the performative prediction literature.
A key concept in performative prediction is the ``distribution map,'' which characterizes how different model deployments impact the population. In convergence analyses this map is typically treated as a fixed and unknown quantity. We study how different types of strategic reasoning impact the distribution map, thus also impacting the resulting convergence properties and equilibria.

A more recent literature on algorithmic collective action~\citep{hardt2023algorithmic} studies coordinated agent efforts with the goal of steering learning systems; see~\citep{baumann2024algorithmic, ben2024role,gauthier2025statisticalcollusioncollectiveslearning,sigg2024decline} for recent developments in this area, as well as related discussions of data leverage~\citep{vincent21dataleverage} and protective optimization technologies~\citep{kulynych20POT}. From the perspective of our work, collective action is a type of look-ahead reasoning: agents plan through model updates under the assumption that they coordinate with other agents. In this work we study the tradeoffs and implications of coordination on the utility of agents participating in the collective. 
Relatedly, \cite{hardt2022power} discuss how platforms can reduce risk by actively steering a population. Collective action reverses this perspective and investigates how the population can improve its utility by steering the learner. This perspective is related to \citep{zrnic2021leads}, who also deviate from the classical model of strategic classification and instead model the population as the leader in the Stackelberg game against the learning platform. Our framework aims to bridge strategic classification and algorithmic collective action to illuminate the benefits and limits of coordination.

Finally, at a technical level, our work leverages ideas from game theory.
\citet{balduzzi2018mechanics} proposed the decomposition of differentiable games into the ``Hamiltonian'' part and the ``potential'' part through the decomposition of the Hessian of the game. Our analysis also relies on the game Hessian, specifically to characterize how the learner’s and the collective’s utilities align or misalign under strategic behavior.


\section{Setup}

We consider a population of individuals interacting with a learning platform.  We assume the platform trains a predictive model on the population's data, and individuals strategically alter their data to achieve favorable outcomes. We elaborate below.

\paragraph{Learning platform.}
Upon observing data about the population, the learner optimizes the parameters $\theta\in\Theta$ of their predictive model $f_\theta$.  We work with the following optimality assumption on the learning algorithm: 
given a loss function $\ell$, the learner's response $\calA(\cD)$ to a data distribution $\cD$ is given by \emph{risk minimization}, defined as
\begin{equation*}\mathcal A(\mathcal D) := \argmin_{\theta\in\Theta}\; \mathbb E_{z\sim \cD}\;[\ell(z,\theta)].\end{equation*}

\paragraph{Strategic agents.} We assume individuals are described by data points $z\in \cZ$ sampled from a base distribution $\cD_0$. Typically, $z = (x,y) \in \cX \times \cY$ are feature--label pairs. Individuals implement a data modification strategy 
\[h_\theta:\cZ\to\cZ\] 
that maps an individual's data point $z$ to a modified data point $h_\theta(z)$; the strategy can depend on the learning platform's currently deployed model $\theta$. We will sometimes omit the subscript $\theta$ if the strategy is independent of the current model, i.e., $h_\theta \equiv h_{\theta'}$ for all $\theta,\theta'$. We use $\cD_{h_\theta}$ to denote the distribution of $h_\theta(z)$ for $ z\sim \cD_0$; in other words, this is the distribution of data points after applying strategy $h_\theta$ to all base data points. Following the terminology of \citet{perdomo20performativeprediction}, we call $\cD_{h_\theta}$ a \emph{distribution map}. Note that different strategies $h_\theta$ correspond to different distribution maps. When the strategy is clear from the context, we will write $\cD_{h_{\theta}} \equiv \cD(\theta)$. 
The notation $\cD_h$ equally applies to model-independent strategies $h$.

\paragraph{Equilibria and learning dynamics.}
We study the long-term behavior of the learner repeatedly optimizing their model and the population strategically adapting to it. Formally, we study the learning dynamics of \emph{repeated risk minimization}:
\begin{equation*}
\label{eq:retraining}
\theta_{t+1} = \calA(\cD_{h_{\theta_t}}).
\end{equation*}
The model $\theta$ is repeatedly updated based on observed data from the previous time step.
The natural equilibrium of these dynamics is called \emph{performative stability}~\citep{perdomo20performativeprediction}.
We say a model $\theta^*$ is performatively stable with respect to a strategy $h_{\theta}$ if
\begin{equation*}\theta^*=\mathcal A(\cD_{h_{\theta^*}}).
\label{eq:PS}
\end{equation*}
In words, there is no reason for the learner to deviate from the current model, given observations of the strategic response of the population.

\paragraph{Population utility.}
Different strategies $h_\theta$ lead to different equilibria. Rather than just focusing on the learner's loss, we evaluate equilibria in terms of the utility they imply for the population.
Let $\theta^*$ denote the equilibrium state under strategy $h_\theta$, i.e., the performatively  stable point. Consequently, $\cD_{h_{\theta^*}}$ denotes the equilibrium distribution. We denote the population's utility after implementing strategy $h_\theta$ by
\begin{equation*}
U(h_\theta)=\expectsub{z\sim\cD_{h_{\theta^*}}}{u(z,\theta^*)},
\end{equation*}
where $u(z,\theta)$ is the utility of an individual with data point $z$ when the deployed model is $\theta$. Note that the stable point satisfies $\theta^*=\mathcal A(\cD_{h_{\theta^*}})$ and thus the utility is evaluated against the data that determines the predictive model. Again, we will sometimes omit the subscript when denoting the strategy if it is independent of the deployed model.


\section{Level-$k$ reasoning}\label{sec: Cog-hier}

Strategic classification \citep{hardt16strategic} assumes that each individual selfishly best-responds to a deployed model $\theta$. As explained earlier, this model does not account for the agents' awareness that they, as a whole, determine the deployed model. To account for this dimension of reasoning, we build on the cognitive hierarchy framework from behavioral economics~\citep{nagel95} and generalize strategic classification to allow individuals to reason through the other individuals' responses. In particular, we formalize \emph{level-$k$} thinking, which categorizes players by the ``depth'' of their strategic thought. Intuitively, an individual reasoning at level $k$ assumes a level of cognitive reasoning for the rest of the population and tries to ``outsmart'' them. In other words, they are always one step ahead: a level-$k$ thinker best-responds to the model that would result from a population of level-$(k-1)$ thinkers.
The basic level-$k$ model starts with an explicit assumption about how individuals at level $0$ behave. It then defines higher levels of thinking recursively.

\newpage
Suppose that agents at level $0$ are non-strategic and implement $h^{(0)}_\theta(z)=z$ in response to all $\theta$. Then, for every higher level of thinking $k\geq 1$ we define the strategy for level-$k$ thinkers recursively as
\begin{equation}
h^{(k)}_{\theta}(z):=\mathrm{argmax}_{z'} \; u\big({z'},\mathcal A(\cD_{k-1}(\theta))\big), \label{eq: def-cog-level}
\end{equation}
where $\cD_{k-1}(\theta)$ is the distribution obtained by applying the model-dependent strategy $h^{(k-1)}_\theta$  to every $z\sim \cD_0$.
At level $k=1$, we recover the standard microfoundation model of strategic classification~\citep{hardt16strategic}, where individuals best-respond to a fixed model.
For larger $k$, the agents anticipate the actions of other agents and best-respond to the hypothetical model resulting from the shifted distribution. The hypothetical model being $\theta'=\mathcal A(\cD_{k-1}(\theta))$.

Different individuals in the population might implement different levels of reasoning.
To reflect this we deviate from a homogeneous population and let the population consists of level-$k$ thinkers at different levels $k$. In particular, we assume that an $\alpha_k$-fraction of the population has cognitive level $k$, for $k=1,2,\dots$ and  $\sum_{k = 1}^\infty \alpha_k=1.$ If $\alpha_k=1$ for some $k$, then all individuals in the population have the same level of reasoning. This model results in the distribution map:
\begin{equation}
\label{eq:level-k-distmap}
{\cD}(\theta) := \sum_{k=1}^\infty \alpha_k \cD_k(\theta).
\end{equation}

We characterize the learning dynamics for different levels of thinking. We use the following Lipschitzness assumption on the induced distribution at level~$k=1$: 
\[
 \calW(\cD_1(\theta), \cD_1(\theta')) \leq \epsilon \norm{\theta - \theta'}, \quad \forall \theta,\theta'\in\Theta,
\]
where $\calW$ denotes the Wasserstein-1 distance. In performative prediction this condition is known as $\epsilon$-sensitivity~\citep{perdomo20performativeprediction}.

\begin{theorem}[Retraining with level-$k$ thinkers]
\label{thm:levelk}
   Suppose the loss of the learner $\ell$ is $\gamma$-strongly convex in $\theta$ and $\beta$-smooth in $z$, and that the distribution map $\cD_{1}(\theta)$ is $\epsilon$-sensitive. Then, as long as  $\epsilon <\frac \gamma \beta$, there is a unique stable point $\theta^*$ such that for any $(\alpha_k)_{k=1}^\infty$ retraining on the mixed population~\eqref{eq:level-k-distmap} converges as 
    \begin{equation*}
    \norm{\theta_t - \theta^*} \leq \left(\sum_{k=1}^{\infty} \left(\frac{\epsilon \beta}{\gamma}\right)^{k} \alpha_{k}\right)^{t}\norm{\theta_0 - \theta^*}.
    \end{equation*}
    \label{thm: cognitive-hierarchy-convergence-general}
\end{theorem}

The core technical step in the proof is to derive how the sensitivity of the distribution map $\cD_k(\theta)$ changes recursively with $k$. In particular, the distribution map $\cD(\theta)$ in~\eqref{eq:level-k-distmap} has sensitivity $\sum_{k=1}^\infty \alpha_k\left(\epsilon\beta/\gamma\right)^{k-1}\epsilon$. We refer to Appendix~\ref{app:proofs} for the full poof.

For the case where $\alpha_1=1$ and thus all agents reason at level $k=1$, we recover the retraining result of \cite{perdomo20performativeprediction}. 
There are two interesting implications of the generalization in Theorem \ref{thm:levelk}. First, we observe that for populations with higher levels of thinking $k$, the rate of convergence increases (although the condition for convergence, $\epsilon<\gamma/\beta$, remains the same). This can be interpreted as saying that performative distribution shifts are mitigated when the population has a deeper level of strategic thought. The second implication is that, as long as the agents act selfishly, they cannot benefit from higher levels of reasoning at stability.

\begin{corollary}
Under the assumptions of Theorem~\ref{thm: cognitive-hierarchy-convergence-general}, it holds that \[U\big(h_\theta^{(1)}\big)=U\big(h_\theta^{(k)}\big),\;\forall k\geq1.\]
Moreover, the utility at stability remains unaltered for any mixed population consisting of level-$k$ thinkers regardless of  $(\alpha_k)_{k=1}^\infty$. 
\label{cor:stable}
\end{corollary}

This corollary follows from the observation that the stable point $\theta^*$ is the same for any mixed population of level-$k$ thinkers. Another consequence of this fact is that the equilibrium strategies $h^{(k)}_{\theta^*}$, and hence the induced distributions $\cD_k(\theta^*)$, are identical for every $k$. 
We will denote this unique \emph{optimal selfish strategy} by $h^*=h^{(k)}_{\theta^*}$ and the implied data distribution by $\cD^*$.


\section{Collective reasoning}
\label{sec:coordinate}
Level-$k$ thinkers anticipate model changes implied by the population's actions. We saw that higher levels of selfish reasoning do not improve the agents' utility at equilibrium. The fundamental reason is that individually they cannot steer the trajectory of the learning algorithm; they can merely anticipate it. In the following we show how individuals can achieve more favorable outcomes by joining forces and making decisions \emph{collectively}; this gives them steering power.

We denote by $h^\sharp$ the \emph{optimal collective strategy}:
\[h^\sharp = \argmax_h U(h) = \argmax_{h} \;\expectsub{z \sim \cD_h }{u(z, \mathcal{A}(\cD_h))}.\]
Notice the difference compared to \eqref{eq: def-cog-level}. In \eqref{eq: def-cog-level}, the optimization variable $z'$ does not enter the model training $\calA$, while above $h$ directly determines the subsequently deployed model. The optimal collective strategy is a Stackelberg equilibrium: the population acts as the Stackelberg leader.

To contrast the optimal collective strategy with the optimal selfish strategy, we define the benefit of coordination.

\begin{definition}[Benefit of coordination] Let $h^*$ be the optimal selfish strategy and $h^\sharp$ the optimal collective strategy. We define the benefit of coordination as 
\[\mathrm{B} = U(h^\sharp)-U(h^*).\]
\end{definition}
Since $h^\sharp$ is the globally optimal strategy for the population, it holds that $\mathrm{B}\geq 0$. 
How large $\mathrm B$ is depends on the goals pursued by the learner and the population, as characterized by $\ell$ and $u$, respectively. Through coordinated data modifications, the population can steer the model towards a common target. But to do so, they may have to deviate from the individually optimal strategy. Thus, what governs the benefit of coordination is the tradeoff between the return of steering the model to a better equilibrium and the cost of being evaluated against the modified data.

We start with a simple case where the benefit of coordination is zero.

\begin{prop} 
Suppose $u = c\cdot \ell$ for some $c\neq 0$. Then, it holds that $\mathrm{B} = 0$.
\label{prop:zero}
\end{prop}

There are two different mechanisms at play, depending on the sign of $c$.
In an adversarial setting where $c > 0$, the game between the learner and the collective is a zero-sum game. In this case the benefit of coordination is zero, as the cost of steering is equal to its return. When $c < 0$, the platform and the agents pursue the same goal and the game becomes a potential game between the two. In this case the benefit of coordination is zero as selfish actions are simultaneously optimal for the collective and there is no benefit to steering the model away from the selfish equilibrium. In general, however, the benefit of coordination can be arbitrarily large. Consider the following example.

\begin{example}[Label modifications as an effective collective lever.]
\label{ex:label}
Consider a learner who aims to accurately predict labels from features and a collective that prefers these predictions to follow a target function $g$. We assume that agents can easily manipulate their labels, while their features cannot be changed. The population and the learner have the following utility and loss, respectively:
\begin{align*}
u(z,\theta)&= - (f_\theta(x)-g(x))^2, \quad \ell(z,\theta)=(f_\theta(x)-y)^2.
\end{align*}
In this setting, the collective has a powerful lever to steer the model through label modifications. The optimal collective strategy is clearly
$h^\sharp(z)=(x,g(x))$. In contrast, selfish agents have no leverage over the learner and would simply report $h^*(z)=(x,y)$. Assuming there exists $\theta$ such that $f_\theta(x) = \mathbb{E}[y|x]$, this gives a benefit of coordination equal to the suboptimality of the labeling function $\mathrm B=\mathbb E_{z\sim \cD_0}(\mathbb{E}[y|x]-g(x))^2$, which can be arbitrarily large, depending on the base data $\cD_0$ and the target function $g$.
\end{example}

To characterize when the benefit of coordination is large, we consider linear distribution maps in the following. A generalized version of the result can be found in Appendix~\ref{app:general-thm}.

\begin{assumption}[Linearity]\label{ass: linearity-of-parameter}
    Let each strategy \(h(\eta)\) be represented by a parameter vector 
    \(\eta \in \mathbb{R}^d\).
    We say that the induced distribution is linear with respect to the parameterization if
    \[
\mathcal{D}_{h(\alpha \eta + (1-\alpha)\eta')}
        = \alpha \mathcal{D}_{h(\eta)}
        + (1-\alpha) \mathcal{D}_{h(\eta')},
        \qquad \forall~\alpha \in [0,1].
    \]
\end{assumption}

Intuitively, linearity means that the population's data distribution is the same whether agents linearly interpolate between two strategies $\eta$ and $\eta'$, or they split up in two subgroups and each implements one of the two strategies. 
Under this assumption, the following result provides a bound on the benefit of coordination.
The proof can be found in Appendix~\ref{app:proofs}.

\begin{theorem}[Bound on the benefit of coordination]
Let \Cref{ass: linearity-of-parameter} hold. Let $U(h(\eta))$ be $\gamma$-strongly concave in $\eta$ and $U(\alpha h  + (1-\alpha) h')$ be differentiable with respect to $\alpha$. Then, we have
    \[
    \mathrm{B} \leq \frac{1}{2\gamma}\Phi^2,
    \]
where \(\quad \Phi:=\innerH{\expectsub{z\sim \cD_{h^*}}{\nabla_\theta u(z, \theta^*)}}{\expectsub{z\sim \cD_{h^\sharp}}{\nabla_\theta \ell(z, \theta^*)}}
    \text{ and } \quad \mathbf{H}^\star =\expectsub{z\in \cD_{h^*}}{\nabla_{\theta,\theta}^2 \ell(z,\theta^*)}.\)
    \label{thm: price-of-selfish-full}
\end{theorem}

This result shows how the benefit of coordination  is governed by the alignment between the utility $u$ of the population and the loss $\ell$ of the learner, quantified by the inner product of their gradients at the selfish equilibrium $\theta^*$. We refer to $\Phi$ as the alignment term. 

We have $\Phi=0$ in the case where the gradients are orthogonal and the two functions $u$ and $\ell$ are unrelated; in this case agents can optimize their utility independent of the model and there is no benefit to being able to steer the model. Further, in line with Proposition~\ref{prop:zero}, when $u = c\cdot \ell$ for some constant $c$, we have $\Phi = 0$ because selfish actions are simultaneously collectively optimal and shifting the model away from $\theta^*$ would not benefit the collective. In particular, $\theta^*$ is a local optimum of the loss under $\cD_{h^*}$ and thus the first term in the inner product becomes zero.

To see cases where $\Phi^2$ can be large, we look at the individual terms in its definition. The first term in the inner product that defines $\Phi$ describes the suboptimality of $\theta^*$ for the agents under $\cD_{h^*}$. It is non-zero if the collective's utility on  $\cD_{h^*}$ can be improved by moving the model away from $\theta^*$. The second term in the inner product captures the response of the learner to the collective strategy. For this term to be non-zero the data modification $\cD_{h^\sharp}$ must make $\theta^*$ suboptimal. Consequently, $\Phi$ is large if both terms in the inner product are non-zero, and there is overlap in the direction of improvement for the learner and the collective. Moreover, what really matters is not just the raw gradient alignment but one filtered through the local curvature of the loss landscape. The directions that the learner finds ``flat'' (small Hessian eigenvalues) allow for more influence on the model through small data modifications, and thus they offer more leverage for the collective. We provide a simple example satisfying the assumptions of Theorem~\ref{thm: price-of-selfish-full} with a non-trivial closed-form expression for $\Phi$ in Appendix~\ref{app:triangle-example}.

Let us revisit Example~\ref{ex:label}. Although the example does not satisfy the assumptions in Theorem~\ref{thm: price-of-selfish-full}, the alignment term provides the right intuition for why $\mathrm B$ is large: since the learner aims to accurately predict labels, any systematic label modification induces a model change. The collective can fully control the direction of these changes, i.e., the gradient of the loss, and align it with their objectives. The label modification strategy leverages this to increase the agents' utility, providing non-zero return as long as $\theta^*$ is suboptimal for the collective under $\cD_{h^*}$.\looseness=-1


\section{Heterogeneous populations} \label{sec: partial-participation}

A perfectly coordinated population where every agent participates, or one that implements the \emph{optimal} collective strategy, is unlikely to emerge in practice.
In the following we consider some plausible deviations from the idealized collective studied in the previous section and discuss how this impacts agent utilities and outcomes. Unless stated otherwise, we assume the collective implements any fixed strategy $h$ (independent of $\theta$), which could be a simpler alternative to a potentially hard-to-implement optimal strategy $h^\sharp$.

\subsection{Scaling strategies in the presence of non-strategic agents}
\label{sec:fixedH}

First, we consider a setting where only a fraction of the population participates in the collective and study conditions under which it is worth scaling up a strategy $h$, meaning a larger collective implies a higher utility for the collective.
To study how the collective's utility changes with its size, we consider the following mixture model for the population as proposed in \citep{hardt2023algorithmic}:
\begin{equation}
\cD^\alpha = \alpha \cD_{h} + (1-\alpha)\cdot  \cD_0
\label{eq: mixture-population-D0}
\end{equation}
where an $\alpha$-fraction of the population implements the collective strategy $h$ and  the remaining $(1-\alpha)$-fraction is non-strategic. Here, the collective strategy $h$ can be any fixed strategy, not necessarily the optimal one. 
In the following, we are interested in the average utility for agents participating in the collective, denoted as $U_\alpha:=\expectsub{z\sim \cD_h}{u(z;\theta^*_\alpha)}$, where $\theta^*_\alpha$ is the equilibrium under the mixture model \eqref{eq: mixture-population-D0}.

\begin{prop}[Benefit of scaling up a strategy]
\label{prop:scaling_up}
Consider the mixture model in~\eqref{eq: mixture-population-D0}, fix a strategy $h$, and denote the resulting equilibrium by $\theta_\alpha^*$. Then, the benefit of scaling up the strategy $h$ for a collective of size $\alpha$ is positive, i.e.,
\[\frac{\partial U_\alpha} {\partial \alpha}> 0, \quad\text{if and only if}\quad
\Psi <
 0,\quad\] 

 \noindent
 with $\Psi=\inner{\expectsub{z \sim \mathcal{D}_{h}}{\nabla_\theta u(z; \theta^*_\alpha)}}{\expectsub{z \sim \mathcal{D}_{h}}{\nabla_\theta \ell (z; \theta^*_\alpha)}}_{\bbH^{-1}}$ where \(\;\bbH := \nabla^2_{\theta}\expectsub{z \sim \mathcal{D}^{\alpha}}{\ell(z;\theta^*_\alpha)}.\)
\end{prop}

The result provides a condition for determining whether a strategy is worth scaling up or not, evaluated with respect to the equilibrium it induces. It is a direct consequence of the envelope theorem and the condition is again linked to a notion of alignment described by $\Psi$. To provide intuition for the result we consider some special cases.
Suppose $\bbH$ is positive semidefinite, which happens when $\ell$ is convex; 
 then, if $u = \ell$, $\Psi \geq 0 $, and if $u = -\ell$, $\Psi\leq 0 $. This means that, under convex losses, when considering the utility of participating agents, scaling up a fixed strategy is always harmful in a zero-sum game, and it is always beneficial when the learner and the population optimize the same target. Note that the result holds for any fixed strategy $h$. We provide additional intuition and empirical insights into how $U_\alpha$  and $\Psi$ change with $\alpha$ in Section~\ref{sec: simulations}. In particular, with a concrete example, we illustrate why larger $\alpha$ can be a disadvantage for the collective, and how the sign change of $\Psi$ accurately characterizes the optimal collective size $\alpha\in(0,1)$ for a fixed strategy~$h$.~\looseness=-1

\paragraph{Optimal size-aware strategy.}
Next, we aim to understand what collectives can achieve if they are aware of partial participation and optimize their strategy accordingly.
We define the optimal size-aware collective strategy for size $\alpha$ as:
\begin{equation*}
 h^\sharp_\alpha = \argmax_{h}\expectsub{z\sim \cD_h}{u(z;\mathcal{A}(\alpha\cD_h + (1-\alpha)\cD_0)}.
\end{equation*}

 The global Stackelberg solution $h^\sharp$ corresponds to the case where $\alpha = 1$ and the collective utility corresponds to the population utility. In the case where $\alpha < 1$,  the collective optimizes the utility of participants, rather than the full population. Agents are informed of their collective size and will choose the best strategy $h^\sharp_\alpha$ accordingly. In the following proposition, we characterize the utility of agents participating in a collective that deploys a size-aware strategy. We use $U^*_\alpha$ to denote the utility of a population of size $\alpha$ implementing the optimal size-aware strategy $h^\sharp_\alpha$.

\begin{prop}[Benefit of larger collectives]
Consider the mixture model~\eqref{eq: mixture-population-D0} with a collective of size $\alpha$ implementing the optimal size-aware strategy $h=h_\alpha^\sharp$. Then, the average collective utility $U^*_\alpha$ achieved by implementing $h^\sharp_\alpha$ satisfies
\[\left. \frac{\partial U^*_\alpha} {\partial \alpha} \geq 0 \quad \text{if and only if}\quad\frac{\partial U_\alpha} {\partial \alpha}\right|_{h=h^\sharp_\alpha}\geq 0.\] \label{prop:benefit-size}
\end{prop}
Note that the derivative in the first term takes into account the dependence of the strategy on $\alpha$. Thus, the result implies that reoptimizing the strategy as a function of collective size does not change whether scaling up is worth it or not. All that matters is the alignment of the pursued goals. The argument involves considering how the equilibrium changes after reoptimizing the strategy for the new size $\alpha$ and evaluating this change against the overall change in the population. A consequence is that, in the case of a zero-sum game and convex losses, the maximum utility a collective \emph{can achieve} decreases with size.

\subsection{Learning dynamics in the presence of selfish agents}

Finally, we study how partial participation impacts learning dynamics  of repeated risk minimization. For $\alpha=1$ and a fixed strategy $h$ that is independent of $\theta$ the learning dynamics converge in a single step. This also holds for $\alpha<1$ in the presence of non-strategic agents. However, this changes as soon as agents deviating from the collective strategy act selfishly. To reflect this scenario we consider the following alternative mixture model:
\begin{equation}
\cD^\alpha(\theta) = \alpha \cD_{h} + (1-\alpha)\cdot  \cD(\theta),
\label{eq: mixture-population}
\end{equation}
where an $\alpha$-fraction of the population implements the collective strategy and an $(1-\alpha)$ fraction deviates from it. We assume these latter act selfishly and their behavior can be characterized by $\cD(\theta)$. In particular, the actions of these agents can depend on the deployed model, such as in level-$k$ reasoning discussed in Section~\ref{sec: Cog-hier}. 

We characterize the rate of convergence of repeated risk minimization under this model and show that larger collectives have the advantage of stabilizing the learning dynamics.
 
\begin{prop}
   Consider the heterogeneous population model~\eqref{eq: mixture-population}. Suppose $\ell$ is $\gamma$-strongly convex in $\theta$ and $\beta$-smooth in $z$, and that the distribution map $ \cD(\theta)$ is $\epsilon$-sensitive. Then, as long as  $\epsilon <\frac \gamma \beta$, repeated risk minimization is guaranteed to converge to a unique stable point $\theta^*_\alpha$ at rate 
    \begin{equation*}
    \norm{\theta_t - \theta^*_\alpha} \leq \left(\frac{\epsilon\beta(1-\alpha)}{\gamma}\right)^{t}\norm{\theta_0 - \theta^*_\alpha}.
    \end{equation*}
    \label{prop:convergence-mixture}
\end{prop}

This result explains how the sensitivity of the non-participating agents to changes in the deployed model, together with the fraction of these agents, determines the rate of convergence to stability. For $\alpha=0$ we again recover the  result of \cite{perdomo20performativeprediction}. The smaller $\alpha$ the slower the rate of convergence. 


\section{Simulations} \label{sec: simulations}

We validate our theoretical findings empirically. We adapt the credit-scoring simulator from \citep{perdomo20performativeprediction} that models how a lending institution classifies loan applicants by creditworthiness.\footnote{For the implementation of the simulation, see \href{https://github.com/haiqingzhu543/Look-Ahead-Reasoning-on-Learning-Platforms}{https://github.com/haiqingzhu543/Look-Ahead-Reasoning-on-Learning-Platforms}; for the dataset, see \href{https://www.kaggle.com/c/GiveMeSomeCredit/data}{https://www.kaggle.com/c/GiveMeSomeCredit}.} We first focus on the results related to level-$k$ reasoning from Section~\ref{sec: Cog-hier} and then offer empirical insights into the trade-offs of collective reasoning from Section~\ref{sec:coordinate} and Section~\ref{sec: partial-participation}. 

\begin{figure}[tb]
  \centering
  \includegraphics[width=0.75\textwidth]{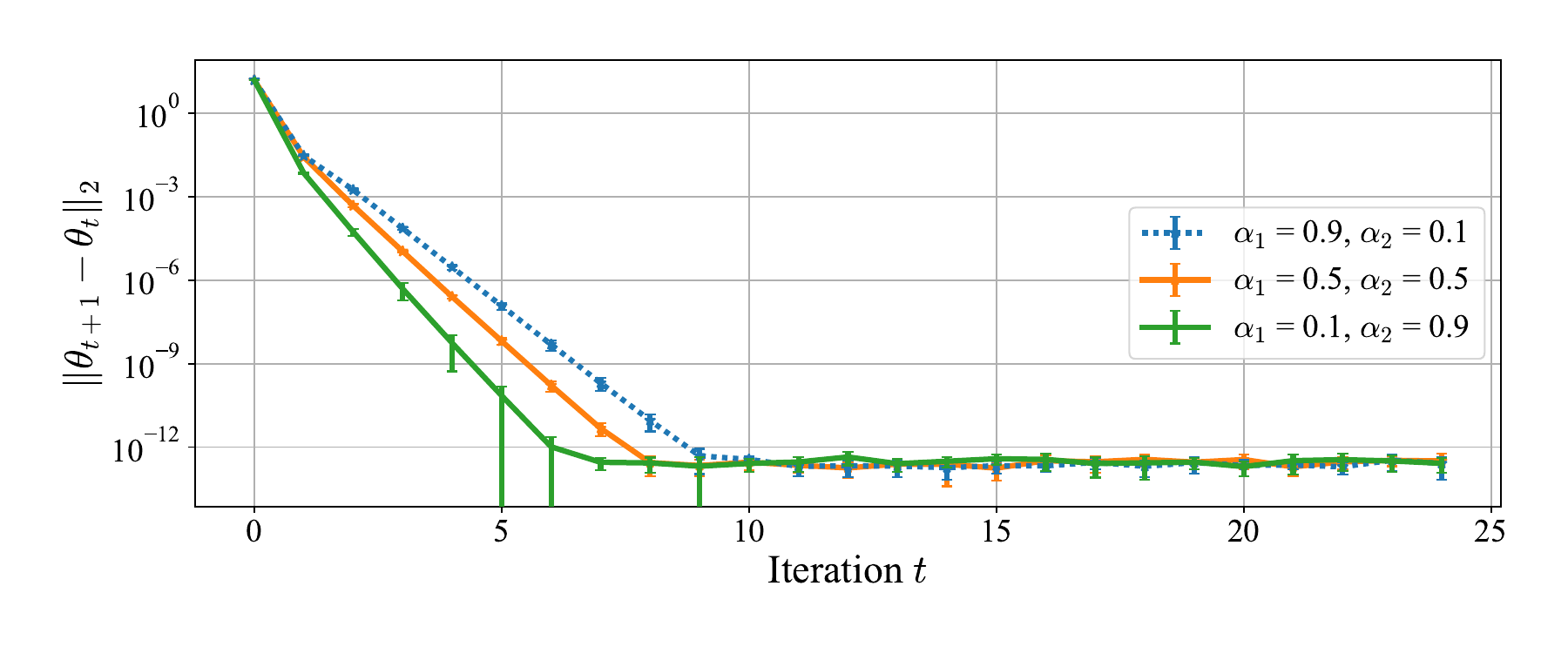}%
  \vspace{-0.2cm}
      \caption{\emph{Convergence of repeated risk minimization on a mixture population of level-$k$ thinkers.} The curves show how the gap between iterates $\norm{\theta_{t+1} - \theta_t}$ evolves across iterations $t$ for different mixture weights. Error bars indicate one standard deviation over 10 runs.}
    \label{fig:cog-level-dynamic}
\end{figure}

\subsection{Retraining dynamics under level-$k$ thinking}\label{subsec: experiment1}

 Assume the learner fits a logistic regression classifier $\theta$ using cross-entropy loss. The data has $10$ features and we assume agents can manipulate the subset $S=\{$`remaining credit card balance',`open
credit lines', `number of real estate loans'$\}$. Given some $\epsilon > 0$, the utility of the agents is given by
\begin{equation*}
   u_\epsilon((x,y), \theta) =  -\inner{\theta}{x} - \frac{1}{2\epsilon}\norm{x_0 - x}^2, \label{eq: utility-simultaion}
\end{equation*}
where $x_0$ is their feature value under $\cD_0$. The best response of the agents is given by $x^*_S = x_S - \epsilon\theta_S$, where $S$ indexes the strategic features. Note that this corresponds to the strategy for agents reasoning at level-1; indeed, assuming other agents are non-strategic implies strategizing against a fixed model. It is not hard to see that the resulting distribution map $\cD_1(\theta)$ is $\epsilon$-sensitive. 
Under this model we simulate the repeated retraining dynamics for mixed populations of level-$1$ and level-$2$ thinkers of varying proportion.

In \Cref{fig:cog-level-dynamic} we report the speed of convergence  by presenting the iterate gap $\norm{\theta_{t+1} - \theta_t}$ against the number of iterations. We choose $\epsilon = 0.5$. First, we can see that under all three mixture weights the gap tends to zero and the dynamics converge. As the fraction of higher levels of thinking increases, the speed of convergence increases, which confirms our theoretical finding in \Cref{thm: cognitive-hierarchy-convergence-general}. 
We also verified empirically that the dynamics converge to a unique equilibrium independent of $\alpha$.

\begin{figure}
  \centering
\begin{minipage}{.48\textwidth}
        \centering
    \includegraphics[width=1\textwidth]{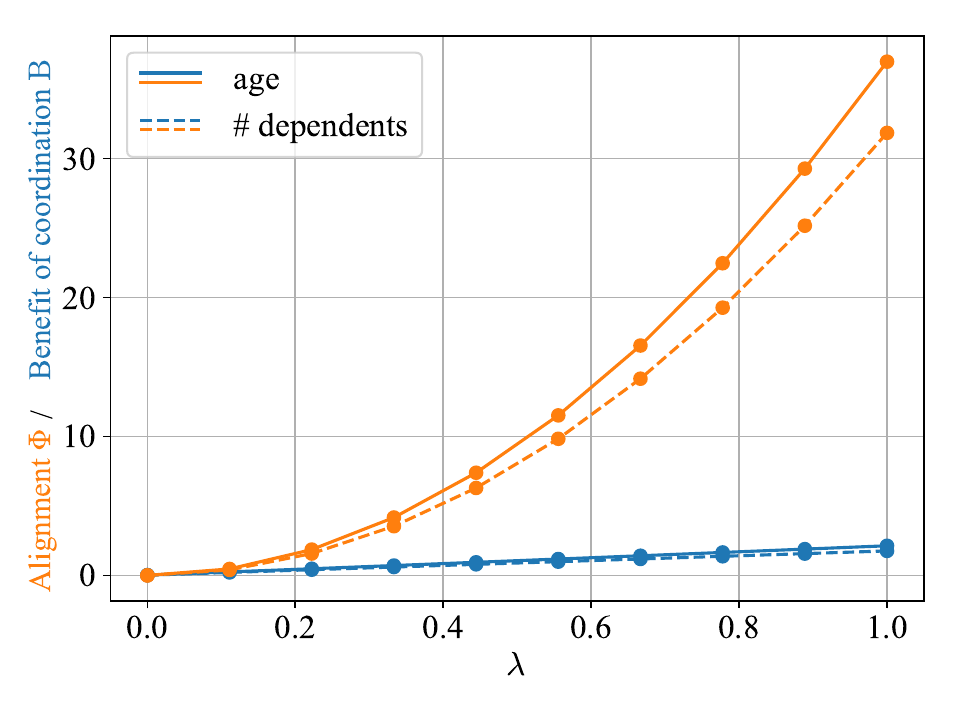}
 \caption{ \textit{Alignment serves as a good proxy for the benefit of coordination.} We consider the utility instantiation in \eqref{eq:util-exp} and evaluate alignment $\Phi$ and the benefit of coordination $\mathrm{B}$ for different values of $\lambda$. We show them for two strategies that modify the feature `age', and `\#dependents', respectively. 
 }\label{fig: alignment-metric-gap-of-util}
  \label{fig:test2}
\end{minipage}
\hfill
\begin{minipage}{.48\textwidth}
  \centering
  \includegraphics[width=1\linewidth]{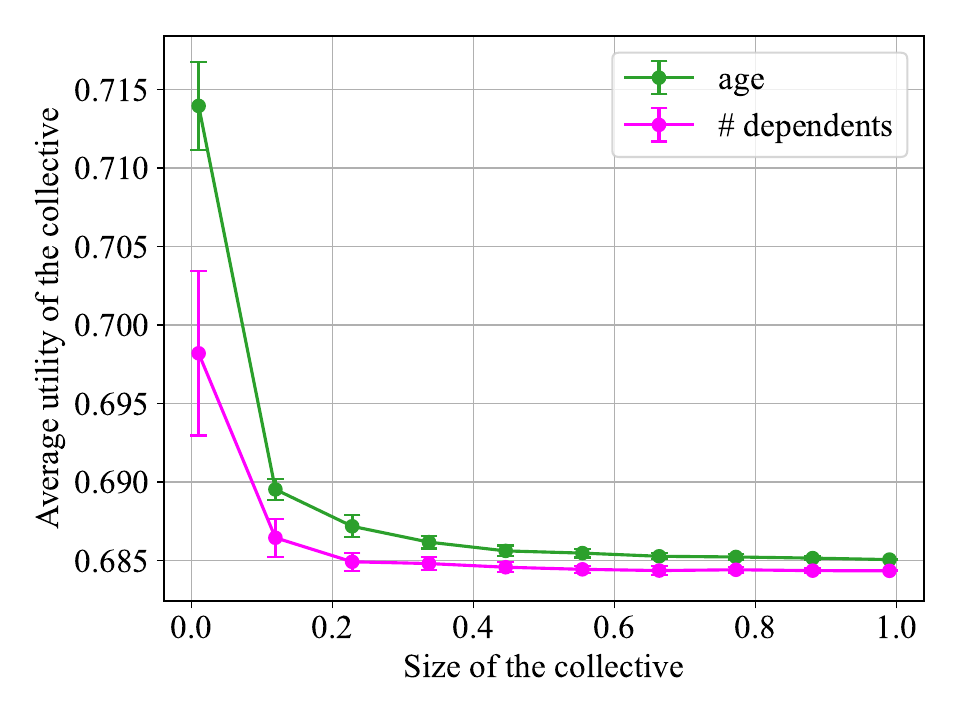}
  \captionof{figure}{\textit{Collective utility decreases with  collective size in the zero-sum case}. The collective implements the optimal size-aware strategy for $\lambda=0$ in a mixed population with non-strategic agents. Small collectives can realize large gains, but the response by the learner impedes gains at larger sizes $\alpha$. 
  }
  \label{fig:optimal-size-strategy}
\end{minipage}
\end{figure}%

\subsection{Trade-offs in collective reasoning}\label{subsec: utilitarian-agents}

With the same credit-scoring data, we study the benefit of coordination and illustrate agent utility for different strategies and collective size. We again consider a learner that trains a logistic regression classifier using cross-entropy loss. For the population, we consider strategies that consist of misreporting a single feature. We choose this to be either `age' or `number of dependents'; `age' is the most important feature for the classification problem, and `number of dependents' is the least important one (see \Cref{fig:misreport} in the Appendix). Contrasting the two strategies allows us to vary the impact of a strategy on the learner in an isolated and systematic way.

\paragraph{Alignment as a proxy for the benefit of coordination.} We investigate empirically how the alignment metric $\Phi$ in Theorem~\ref{thm: price-of-selfish-full} relates to the benefit of coordination.  We instantiate the agents' utility as~\looseness=-1
\begin{equation}
 u\big((x,y);\theta\big) = \ell((x,y),\theta) - \lambda\cdot \Vert \theta \Vert^2, \label{eq:util-exp}
\end{equation}
where $\ell$ is the loss of the learner and the regularization term $\lambda\in[0,1]$ controls the alignment between the learner's and the agents' objectives.  For $\lambda=0$ we get a zero-sum game. The larger $\lambda$ the more different the learner's and the agents' objectives are. 

Note that, in this setting, the linearity and the strong concavity of our theory are not satisfied. We are still interested in investigating to what extent the alignment metric serves as a useful proxy for the benefit of coordination. In Figure~\ref{fig: alignment-metric-gap-of-util} use $\lambda$ to vary alignment and plot $\Phi$ against the benefit of coordination $\mathrm B$. We find that $\mathrm B<\Phi$, which is in line with our theory. Furthermore, alignment correlates with the benefit of coordination and accurately predicts that one strategy is more effective than the other. Recall that we restrict the strategy space to the modification of a single feature; the solid and the dashed line compare two different settings.

\paragraph{Cost of steering and why larger collectives can be worse off.} 
Next, we illustrate the challenges that come with large collectives in a misaligned setup. For this purpose we consider the zero-sum setting with $\lambda=0$
and a mixed-population composed of a collective and non-strategic agents. The collective implements the optimal size-aware strategy $h^\sharp_\alpha$ for modifying each of the two features. We approximate this optimal strategy using gradient descent with learning rate $0.01$ and $250$ epochs.

In Figure~\ref{fig:optimal-size-strategy} we visualize the collective utility as a function of the collective size $\alpha$.  We see that the collective utility is maximized as $\alpha\rightarrow 0$ and decreases with size. The gap at $\alpha=0$ shows the benefit of strategic data reporting against a fixed model. Small $\alpha$ is an advantage in the zero-sum case as agents have diminishing influence on the learner and they can move almost independently of $\theta$. For larger collectives it becomes increasingly hard to realize a benefit because the model responds to the agents' actions and an equivalent change to the feature has a larger effect on the learner's loss. This counter-force in the case of conflicting utilities is the reason why the collective utility may decrease with the collective size.

\begin{figure}
  \begin{subfigure}{0.5\textwidth}
        \centering
    \includegraphics[width=0.95\linewidth]{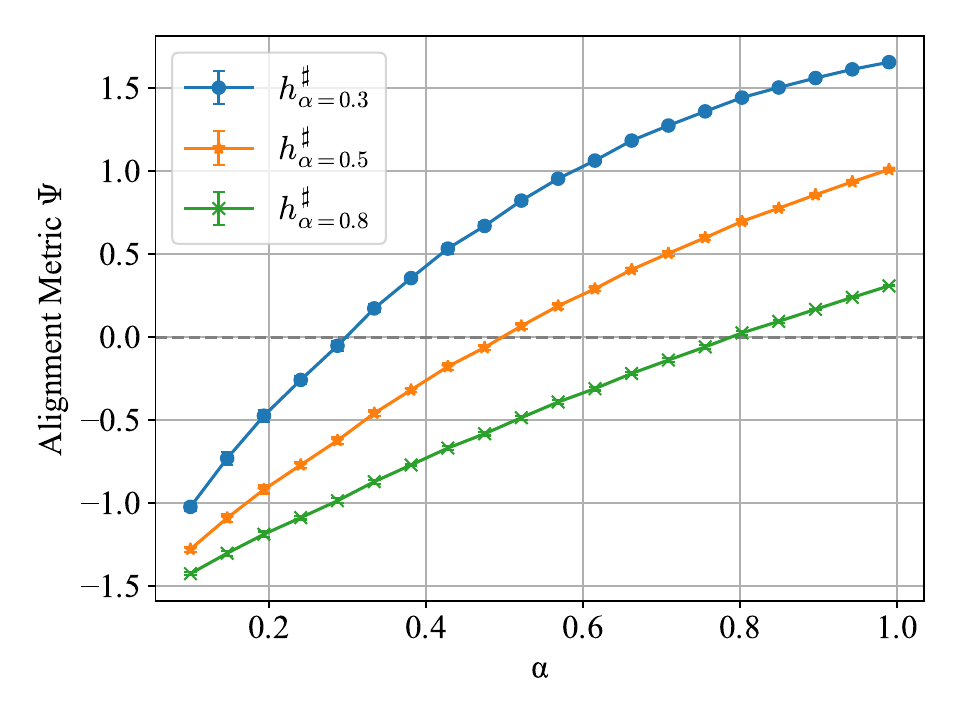}
    \label{fig:fixed-strategy-alignment-metric}
  \end{subfigure}
  \begin{subfigure}{0.5\textwidth}
        \centering
    \includegraphics[width=0.95\linewidth]{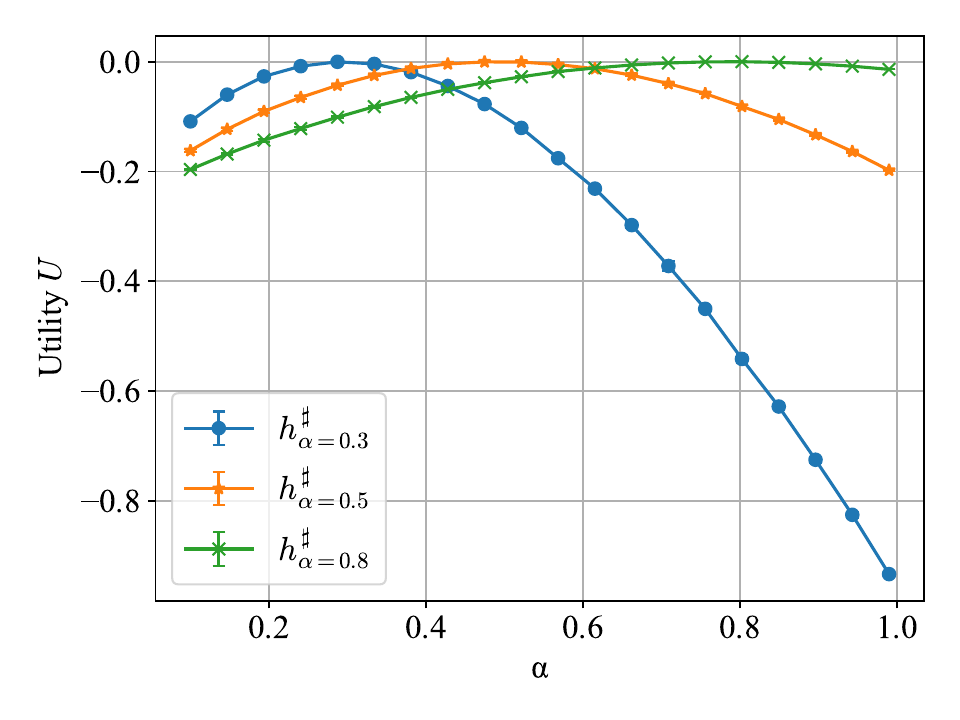}
    \label{fig:fixed-strategy-utility}
  \end{subfigure}
\caption{\emph{Change in alignment metric (left) and utility (right) with collective size, for three fixed strategies.} The utility is non-monotonic in size, and the sign of the alignment metric $\Psi$ accurately predicts whether it is worth scaling up a strategy or not. We consider the setting in \eqref{eq:u-target} and evaluate three different strategies, corresponding to the optimal size-aware strategy $h^\sharp_\alpha$ at $\alpha\in\{0.3,0.5,0.8\}$.}
  \label{fig:fixed-strategy}
\end{figure}

\paragraph{Scaling up a fixed strategy.}
\label{sec:non-monotone}

Finally, we verify Proposition~\ref{prop:scaling_up} empirically. For simplicity, we assume the learner performs binary classification to predict whether a person experienced 90 days past due delinquency or worse. The collective is uniformly sampled from the data with label $1$ and aims to maximize their utility 
\begin{equation}
 u_i((x_i,y_i),\theta) = -\norm{\theta - \theta_{\text{target}}}^2,
 \label{eq:u-target}
\end{equation}
where $\theta_{\text{target}}$ a fixed target model that the collective would like to achieve (see Appendix \ref{app:binary} for details). We again assume the collective can modify individual features and consider three strategies $h^\sharp_{\alpha=0.3}$, $h^\sharp_{\alpha=0.5}$, and $h^\sharp_{\alpha=0.8}$, where each strategy is optimal for a given collective size $\alpha \in \{0.3, 0.5, 0.8\}$. We refer to Appendix~\ref{app:binary} for details on how the strategies are computed. 

In Figure~\ref{fig:fixed-strategy} we illustrate the collective  utility and the alignment metric $\Psi$ as defined in Proposition~\ref{prop:scaling_up} for different collective sizes $\alpha$. We observe that, in accordance with our theory, when the alignment metric is positive, the average utilities of the collective decrease as the sizes of the collective increase. Analogously, when the alignment is negative, the average utilities of the collective increase with collective size. Notably, as $\alpha$ approaches the assumed sizes $0.3, 0.5$ and $0.8$, the utility converges toward $0$, indicating that the model closely matches the target $\theta_{\text{target}}$. Beyond this point, however, the collective ``over-pushes'' the model, leading to a decrease in utility and a positive $\Psi$.

\section{Conclusion}

We study look-ahead reasoning as a new aspect of strategic reasoning on learning platforms. 
While traditional analyses of strategic classification treat users as reacting independently to a fixed model, look-ahead reasoning highlights that users’ incentives and actions are inherently interdependent---each agent’s actions influence future model deployments, and thus the utility of other agents in the population.
Within this broad theme, we find that higher-order reasoning accelerates convergence toward equilibrium but does not improve individuals' long-run outcomes, 
suggesting that attempts to ``outsmart'' others may offer only transient advantages. In contrast, collective reasoning---where users coordinate their behavior through their shared impact on the model---allows the agents to steer the model towards a desirable state. Our results show that this can be a very effective lever for the collective when the loss of the learner and the utility of the population are appropriately aligned. However, for conflicting objectives, we find that the excessive steering power that comes with larger collectives can prevent large utility gains.

A central goal of our work was to provide a unifying framework to contrast selfish reasoning with collective reasoning when contesting machine learning predictions. There are several natural extensions of our work. One is investigating look-ahead reasoning under imperfect information. As common in economic models, we assumed agents determine their actions under perfect information. In our case, this concerns knowledge about the learner's loss function and the population's data distribution. However, in practice this information needs to be estimated from finite data which raises questions of statistical complexity.
A related question would be to study how model misspecifications, estimation errors, and imperfect coordination impact outcomes in look-ahead reasoning.

More broadly, we hope our work can support recent discussions concerning data poisoning and multi-party adversarial attacks~\citep[e.g.,][]{nestaas2025adversarial}, and 
open up new pathways to transfer insights from strategic classification to algorithmic collective action, and vice versa.

\section*{Acknowledgements}
Celestine Mendler-Dünner acknowledges the financial support of the Hector Foundation. The work was conducted during an internship of Haiqing Zhu at the Max Planck Institute for Intelligent Systems, Tübingen.

\bibliographystyle{plainnat}
\bibliography{refs}

\newpage
\appendix

\section{Proofs}

\label{app:proofs}

\subsection{Auxiliary results}
\label{app:general-thm}

\begin{lemma}
    Let $\calW$ denote the Wasserstein-1 distance, and $\sum_{i=1}^n \alpha_i \geq 0$ with $\alpha_i \geq 0$, then
    \[
    \calW\left(\sum_{i=1}^n \alpha_i \cD_i, \sum_{i=1}^n \alpha_i \cD'_i\right) \leq \sum_{i=1}^n \alpha_i \calW(\cD_i,\cD'_i).
    \] \label{app-lem: decomp-wasser-distance}
\end{lemma}
\begin{proof}
    By definition, the Wasserstein-1 distance can be written as
    \[
    \min_{\mu(X,Y)}\expectsub{\mu(X,Y)}{\|X - Y\|},
    \]
    where $\mu$ is the joint distribution of $X,Y$ and $X \sim \sum_{i=1}^n \alpha_i \cD_i$, $Y\sim \sum_{i=1}^n \alpha_i \cD'_i$. Consider the measures $\mu_i$ defined as
    \[
    \mu_i = \min_{\mu(X_i,Y_i)}\expectsub{\mu(X_i,Y_i)}{\|X_i - Y_i\|},
    \]
    where $X_i \sim \cD_i$, $Y_i \sim \cD_i'$. Then, with $\hat{\mu} = \sum_i \alpha_i\mu_i$, we can notice that
    \[
    \calW\left(\sum_{i=1}^n \alpha_i \cD_i, \sum_{i=1}^n \alpha_i \cD'_i\right) \leq \expectsub{\hat{\mu}}{\|X- Y\|} = \sum_{i=1}^n \alpha_i \expectsub{\mu_i}{\|X- Y\|} = \sum_{i=1}^n \alpha_i \calW(\cD_i,\cD'_i),
    \]
    where the first inequality follows from the minimization property of the Wasserstein-1 distance.
\end{proof}


Below we state a bound on the benefit of coordination that does not assume linearity  (Assumption~\ref{ass: linearity-of-parameter}).

\begin{theoremp}{\ref{thm: price-of-selfish-full}'}
    Let each strategy \(h(\eta)\) be represented by a parameter vector 
    \(\eta \in \mathbb{R}^d\). Suppose $U(h(\eta))$ is $\gamma$-strongly concave on $\eta$. Then, we have
    \begin{equation*}
     0\leq \mathrm{B} \leq  \frac{1}{2\gamma}\cdot\norm{  \nabla_\eta \nabla_\theta\expectsub{z\sim \cD_{h^*}}{\ell(z, \theta^*)}(\mathbf{H}^\star)^{-1} \expectsub{z \sim \cD_{h^*}}{\nabla_\theta u(z, \theta^*)}}^2. \label{app-eq: gap-of-util}
    \end{equation*}
    with $\mathbf{H}^\star:= \expectsub{z\in \cD_{h^*}}{\nabla_{\theta,\theta}^2 \ell(z,\theta^*)}$ 
    and $\theta^*$ denoting the stable point corresponding to the selfish strategy $h^*$. 
    \label{thm: PA}
\end{theoremp}


\subsection{Proof of Theorem~\ref{thm: cognitive-hierarchy-convergence-general}}

The key step is to prove Lemma~\ref{lem: sensitivity-mixed} below. 
The claim in the theorem follows by combining Lemma~\ref{lem: sensitivity-mixed} with Theorem 3.5 in \citet{perdomo20performativeprediction}.

\begin{lemma}
 Let $\alpha_k$ be the portion of the population with cognitive level $k$. Suppose $\cD_1(\theta)$ is $\epsilon$-sensitive and the loss $\ell$ is $\gamma$-strongly convex and $\beta$-smooth in $z,\theta$. Then  the distribution map  $\overline{\cD}(\theta) := \sum_{k=1}^\infty \alpha_k \cD_k(\theta)$ is $\bar\epsilon$-sensitive with \[\bar \epsilon = \sum_{k=1}^\infty \alpha_k\left(\frac{\epsilon\beta}{\gamma}\right)^{k-1}\epsilon.\] \label{lem: sensitivity-mixed}
\end{lemma}
\begin{proof}
    Denote $\theta^k := \calA(\cD_1(\theta^{k-1}))$ and $\theta^0 = \theta$. Similarly, $\phi^k := \calA(\cD_1(\phi^{k-1}))$ and $\phi^0 = \phi$. From \Cref{eq: def-cog-level}, we can see that $\cD_{k}(\theta^0) = \cD_1(\theta^{k-1})$. Consider the map $\cD_k$; then, we have the recursion:
    \[
    \calW(\cD_k(\theta), \cD_k(\phi)) = \calW(\cD_1(\theta^{k-1}), \cD_1(\phi^{k-1})) \leq \epsilon \norm{\theta^{k-1} - \phi^{k-1}}.
    \]
    By \citet{perdomo20performativeprediction}, Theorem 3.5, we also have \[\norm{\theta^{k-1} - \phi^{k-1}} \leq \left(\frac{\epsilon\beta}{\gamma}\right)^{k-1} \norm{\theta - \phi} .\] Finally, we notice that
    \begin{align*}
    \calW\left(\sum_{k=1}^\infty \alpha_k \cD_k(\theta), \sum_{k=1}^\infty \alpha_k \cD_k(\phi)\right) &\leq \sum_{k=1}^\infty \alpha_k\calW(\cD_k(\theta), \cD_k(\phi))\leq \sum_{k=1}^\infty \alpha_k \left(\frac{\epsilon\beta}{\gamma}\right)^{k-1}\epsilon \norm{\theta - \phi},
    \end{align*}
    where the first inequality follows from \Cref{app-lem: decomp-wasser-distance}. 
\end{proof}

\subsection{Proof of Corollary~\ref{cor:stable}}

We start from Theorem~\ref{thm: cognitive-hierarchy-convergence-general}. Define the contraction factor
\[\rho_\alpha = \left(\sum_{k=1}^{\infty} \left(\frac{\epsilon \beta}{\gamma}\right)^{k} \alpha_{k}\right).\]
It can be seen that if $\rho_\alpha$ is positive for some $\alpha$, it holds that the dynamics will converge to the equilibrium for any $\alpha$ with $\sum_{k=1}^\infty \alpha_k=1$. Similarly, if it is zero, this holds so for any $\alpha$. Thus, a simple contraction argument shows that the trajectory converges to the same stable point independent of $\alpha$. The same holds for the special case  $\alpha_k = 1$. At this point, no agent is moving and thus the equilibrium strategies $h^{(k)}_{\theta^*}$ are identical, and so are the utilities:
\[h^{(k)}_{\theta^*}=h^{(k')}_{\theta^*}\quad \Rightarrow\quad U(h^{(k)}_{\theta^*})=U(h^{(k')}_{\theta^*}).\]

\subsection{Proof of Proposition~\ref{prop:zero}}
Since $(h^*, \theta^*)$ is the performatively stable point, by definition, it is clear that
\[
   \expectsub{z\sim \cD_{h^*}}{\nabla_\theta\ell(z, \theta^*)} = 0.
\]
Therefore, since $u = c \cdot \ell $, we can conclude that $\expectsub{z\sim \cD_{h^*}}{\nabla_\theta u(z, \theta^*)} = c\cdot\expectsub{z\sim \cD_{h^*}}{\nabla_\theta\ell(z, \theta^*)} =0 $. Then, by \Cref{thm: PA}, we have $\mathrm{B} = 0$.

\subsection{Proof of \Cref{thm: price-of-selfish-full}}
  We use the parameterization in \Cref{ass: linearity-of-parameter}. The optimal strategy is defined as $h^\sharp:= h(\eta^\sharp)$ where $\eta^\sharp := \argmax_{\eta} U(h(\eta))$ . The performatively stable point $(h^*, \theta^*)= (h(\eta^*), \theta^*)$ is the point such that
  \begin{align*}
     \eta^* &= \argmax_{\eta} \expectsub{z\sim \cD_{h(\eta)}}{u(z, \theta^*)}, \\
     \theta^* &= \argmin_\theta \expectsub{z\sim \cD_{h(\eta^\star)}}{\ell(z, \theta)}.
  \end{align*}
  We consider a hypothetical mixture population which is represented as $\alpha \cD_{h^\sharp} + (1-\alpha)\cD_{h^*} = \cD_{h(\alpha \eta^\sharp + (1-\alpha)\eta^*)}$. For this mixture population, $\alpha=0$ indicates the equilibrium of selfish actions. This means that, if we fix the learner's action to be $\theta^*$, the action $h_\alpha = h(\alpha \eta^\sharp + (1-\alpha)\eta^*)$ will maximize the population's expected utility only when $\alpha = 0$. Formally, consider the function $\iota(\alpha) = \expectsub{z\sim \cD_{\alpha h^\sharp + (1-\alpha) h^*}}{u(z,\theta^*)} = \expectsub{z\sim \cD_{h(\alpha \eta^\sharp + (1-\alpha)\eta^*)}}{u(z,\theta^*)}$. By \Cref{ass: linearity-of-parameter}, we must have
    \[
    \frac{\partial \iota}{\partial \alpha}\Big|_{\alpha = 0} = \expectsub{z\sim \cD_{h^\sharp}}{ u(z, \theta^*)} - \expectsub{z\sim \cD_{h^*}}{ u(z, \theta^*)} = 0.
    \]
    At the stable point $(h^*, \theta^*)$, we also have that $ \expectsub{z\sim \cD_{h^*}}{\nabla_\theta \ell(z, \theta^*)} = 0$.
    Consider the function $e(\alpha) = U\left(\alpha h^\sharp + (1-\alpha)h^*\right) = U\left(h(\alpha \eta^\sharp + (1-\alpha)\eta^*)\right)$; we have
    \begin{align*}
    \frac{\partial e}{\partial \alpha}\Big|_{\alpha = 0} &= \expectsub{z\sim \cD_{h^\sharp}}{ u(z, \theta^*)} - \expectsub{z\sim \cD_{h^*}}{ u(z, \theta^*)} \\
    &~~~~~- \innerH{\expectsub{z\sim \cD_{h^*}}{\nabla_\theta u(z, \theta^*)}}{\expectsub{z\sim \cD_{h^\sharp}}{\nabla_\theta \ell(z, \theta^*)}- \expectsub{z\sim \cD_{h^*}}{\nabla_\theta \ell(z, \theta^*)}} \\
    &= -\innerH{\expectsub{z\sim \cD_{h^*}}{\nabla_\theta u(z, \theta^*)}}{\expectsub{z\sim \cD_{h^\sharp}}{\nabla_\theta \ell(z, \theta^*)}- \expectsub{z\sim \cD_{h^*}}{\nabla_\theta \ell(z, \theta^*)}}\\
    &= -\innerH{\expectsub{z\sim \cD_{h^*}}{\nabla_\theta u(z, \theta^*)}}{\expectsub{z\sim \cD_{h^\sharp}}{\nabla_\theta \ell(z, \theta^*)}}.
    \end{align*}
    Finally, the result follows from the PL-inequality.

\subsection{Proof of \Cref{thm: PA}}
\label{app:general-thm}

   For notational simplicity, we set $f(\eta,\theta):=\expectsub{z\sim\cD_{h(\eta)}}{u(z;\theta)}$ and $g(\eta,\theta):=\expectsub{z\sim\cD_{h(\eta)}}{\ell(z;\theta)}$. 
    Recall that
    \[
    U(h(\eta)) = f(\eta, \calA(\cD_{h(\eta)})) = \expectsub{z \sim \cD_{h(\eta)}}{u(z, \calA(\cD_{h(\eta)}))},
    \]
    where the first argument in $f$ only applies in distribution that is taken against and the second argument is corresponding to the second argument of $u$. Then, by the implicit function theorem we have
    \[
    \nabla_\eta U(h(\eta)) =\nabla_\eta f(\eta, \calA(\cD_{h(\eta)}))= \nabla_1 f(\eta, \calA(\cD_{h(\eta)})) - \nabla^2_{1,2} g(\eta,\calA(\cD_{h(\eta)})) \left[\nabla^2_{2,2} g(\eta,\calA(\cD_{h(\eta)}))\right]^{-1}\nabla_2 f(\eta,\calA(\cD_{h(\eta)})),
    \]
    where $\nabla_1$ and $\nabla_2$ denote the gradient operator on the first/second argument of the function. For the equilibrium of the selfish strategy $(h(\eta^*), \theta^*)$, we must have $\nabla_1 f(\eta, \calA(\cD_h)) = 0$. Let $\eta^\sharp := \argmax_\eta U(h(\eta))$, by the PL-inequality, we obtain that
    \[
    U\left(h(\eta^\sharp)\right) - U\left(h(\eta^*)\right) \leq \frac{1}{2\gamma}\cdot \norm{ \nabla^2_{1,2} g(\eta^*,\calA(\cD_{h(\eta^*)})) \left[\nabla^2_{2,2} g(\eta^*,\calA(\cD_{h(\eta^*)}))\right]^{-1}\nabla_2 f(\eta^*,\calA(\cD_{h(\eta^*)}))}^2.
    \]

\subsection{Proof of \Cref{prop:scaling_up}}
Consider the derivative of $ U_\alpha$ with respect to variable $\alpha$,
\[
\frac{\partial U_\alpha}{\partial \alpha} = \frac{\partial \expectsub{z \sim \mathcal{D}_h}{u(z;\theta_\alpha^*)}}{\partial \alpha} = \frac{\partial \expectsub{z \sim \mathcal{D}_h}{u(z;\theta_\alpha^*)}}{\partial \theta} \cdot \frac{\partial \theta^*_\alpha}{\partial \alpha}.
\]
Next, we notice that
\[
\nabla_\theta \expectsub{z \sim \alpha \mathcal{D}_h + (1-\alpha) \mathcal{D}_0}{\ell(z;\theta)} = 0,
\]
where we can further write the left-hand side as
\begin{equation*}
\nabla_\theta \expectsub{z \sim \alpha \mathcal{D}_h + (1-\alpha) \mathcal{D}_0}{\ell(z;\theta)} = \alpha \cdot \expectsub{ z \sim \mathcal{D}_h}{\nabla_\theta \ell(z; \theta)} + (1-\alpha) \cdot \expectsub{ z \sim \mathcal{D}_0}{\nabla_\theta \ell(z; \theta)} = 0.
\end{equation*}
Therefore, $\expectsub{ z \sim \mathcal{D}_0}{\nabla_\theta \ell(z; \theta)} = -\frac{\alpha}{1-\alpha} \cdot \expectsub{ z \sim \mathcal{D}_h}{\nabla_\theta \ell(z; \theta)}$.
Then, we consider the term $\frac{\partial \theta^*_\alpha}{\partial \alpha}$. By the implicit function theorem, with $\alpha > 0$, we have
\begin{align*}
\frac{\partial \theta^*_\alpha}{\partial \alpha} &= - \left(\nabla^2_{\theta, \theta}\expectsub{z \sim \alpha \mathcal{D}_h + (1-\alpha) \mathcal{D}_0}{\ell(z;\theta)}\right)^{-1}\left(\expectsub{ z \sim \mathcal{D}_h}{\nabla_\theta \ell(z; \theta)} -  \expectsub{ z \sim \mathcal{D}_0}{\nabla_\theta \ell(z; \theta)}\right)\\
& = - \frac{1}{1-\alpha}\left(\nabla^2_{\theta, \theta}\expectsub{z \sim \alpha \mathcal{D}_h + (1-\alpha) \mathcal{D}_0}{\ell(z;\theta)}\right)^{-1}\left(\expectsub{ z \sim \mathcal{D}_h}{\nabla_\theta \ell(z; \theta)} \right).
\end{align*}
Therefore, we can finally write
\[
\frac{\partial U_\alpha}{\partial \alpha} = - \frac{1}{1-\alpha}\left(\expectsub{z \sim \mathcal{D}_h}{\nabla_\theta u(z; \theta^*_\alpha)}\right)^T \left(\nabla^2_{\theta, \theta}\expectsub{z \sim \alpha \mathcal{D}_h + (1-\alpha) \mathcal{D}_0}{\ell(z;\theta)}\right)^{-1}\left(\expectsub{ z \sim \mathcal{D}_h}{\nabla_\theta \ell(z; \theta)} \right).
\]

\subsection{Proof of \Cref{prop:benefit-size}}
The proof is the same as the proof of \Cref{prop:scaling_up} up to a use of the envelope theorem. For completeness, we restate the proof here and highlight the use of the envelope theorem.
For notational simplicity, we abbreviate $h^\sharp_\alpha$ as $h$. Consider the derivative of $U_\alpha^*$ with respect to $\alpha$,
\[
\frac{\partial U_\alpha^*}{\partial \alpha} = \frac{\partial \expectsub{z \sim \mathcal{D}_h}{u(z;\theta_\alpha^*)}}{\partial \alpha} = \frac{\partial \expectsub{z \sim \mathcal{D}_h}{u(z;\theta_\alpha^*)}}{\partial \theta} \cdot \frac{\partial \theta^*_\alpha}{\partial \alpha},
\]
where the second equality follows from the implicit function theorem and the envelope theorem. Next, we notice that
\[
\nabla_\theta \expectsub{z \sim \alpha \mathcal{D}_h + (1-\alpha) \mathcal{D}_0}{\ell(z;\theta)} = 0,
\]
where we can further write the left-hand side as
\begin{equation*}
\nabla_\theta \expectsub{z \sim \alpha \mathcal{D}_h + (1-\alpha) \mathcal{D}_0}{\ell(z;\theta)} = \alpha \cdot \expectsub{ z \sim \mathcal{D}_h}{\nabla_\theta \ell(z; \theta)} + (1-\alpha) \cdot \expectsub{ z \sim \mathcal{D}_0}{\nabla_\theta \ell(z; \theta)} = 0.
\end{equation*}
Therefore, $\expectsub{ z \sim \mathcal{D}_0}{\nabla_\theta \ell(z; \theta)} = -\frac{\alpha}{1-\alpha} \cdot \expectsub{ z \sim \mathcal{D}_h}{\nabla_\theta \ell(z; \theta)}$.
Consider the term $\frac{\partial \theta^*_\alpha}{\partial \alpha}$. By the implicit function theorem, with $\alpha > 0$, we have
\begin{align*}
\frac{\partial \theta^*_\alpha}{\partial \alpha} &= - \left(\nabla^2_{\theta, \theta}\expectsub{z \sim \alpha \mathcal{D}_h + (1-\alpha) \mathcal{D}_0}{\ell(z;\theta)}\right)^{-1}\left(\expectsub{ z \sim \mathcal{D}_h}{\nabla_\theta \ell(z; \theta)} -  \expectsub{ z \sim \mathcal{D}_0}{\nabla_\theta \ell(z; \theta)}\right)\\
& = - \frac{1}{1-\alpha}\left(\nabla^2_{\theta, \theta}\expectsub{z \sim \alpha \mathcal{D}_h + (1-\alpha) \mathcal{D}_0}{\ell(z;\theta)}\right)^{-1}\left(\expectsub{ z \sim \mathcal{D}_h}{\nabla_\theta \ell(z; \theta)} \right).
\end{align*}
We finally write
\[
\frac{\partial U_\alpha^*}{\partial \alpha} = - \frac{1}{1-\alpha}\left(\expectsub{z \sim \mathcal{D}_h}{\nabla_\theta u(z; \theta^*_\alpha)}\right)^T \left(\nabla^2_{\theta, \theta}\expectsub{z \sim \alpha \mathcal{D}_h + (1-\alpha) \mathcal{D}_0}{\ell(z;\theta)}\right)^{-1}\left(\expectsub{ z \sim \mathcal{D}_h}{\nabla_\theta \ell(z; \theta)} \right).
\]

\subsection{Proof of \Cref{prop:convergence-mixture}}
By \Cref{lem: sensitivity-mixed}, the sensitivity of the mixture distribution can be computed as
\[
  \sum_{k=1}^\infty \alpha_k \left(\frac{\epsilon \beta}{\gamma}\right)^{k-1} \epsilon = (1-\alpha)\epsilon,
\]
where $\alpha_1 = 1-\alpha$ and $\alpha_0 = \alpha$.
Combining this with Theorem 3.5 in \citet{perdomo20performativeprediction} yields the result.

\section{Example to illustrate Theorem~\ref{thm: price-of-selfish-full}}
\label{app:triangle-example}
\begin{wrapfigure}{r}{0.3\textwidth}
   \centering
  \begin{tikzpicture}[scale=1.4, thick]

    \coordinate (P1) at (-0.866, -0.5);   
    \coordinate (P2) at (0.866, -0.5);    
    \coordinate (P3) at (0, 1);           
    
    \coordinate (ThetaStar)  at (0, 0);
    \coordinate (ThetaSharp) at ($ (P1)!0.5!(P2) $);         
    
    \draw[thick] (P1)--(P2)--(P3)--cycle;
    
    \filldraw[black] (P1) circle (0.6pt) node[below left] {$\mathbf{p}_1$};
    \filldraw[black] (P2) circle (0.6pt) node[below right] {$\mathbf{p}_2$};
    \filldraw[black] (P3) circle (0.6pt) node[above] {$\mathbf{p}_3$};
    
    \filldraw[DarkGreen] (ThetaStar) circle (0.8pt) node[below=2pt] {$\theta^*$};
    \filldraw[DarkRed] (ThetaSharp) circle (0.8pt) node[below=2pt] {$\theta^\sharp$};

    \end{tikzpicture}
\end{wrapfigure}
Consider a toy setting to derive a closed-form expression for the benefit of coordination. Assume the learner estimates the centroid $\theta$ of a distribution  supported on three anchor points ${\mathbf{p}_1, \mathbf{p}_2, \mathbf{p}_3}$—the corners of an equilateral triangle.
The collective applies a strategy $h$ that moves their data point $z$ to one of the three anchors with probability $(w_1,w_2,w_3)$, respectively, $\sum_i w_i = 1$ and $w_i \ge 0$. This implies a distribution $\mathcal{D}_h=\sum_i w_i \delta_{\mathbf{p}_i}$  where the strategy determines $\mathbf w$. The learner minimizes the squared loss; the solution is $\calA(\cD_h) = \sum_i w_i x_i$.
The collective, on the other hand, prefers the centroid to lie between $\mathbf p_1$ and $\mathbf p_2$, and  maximizes
\[u(z, \theta) = -\Vert \mathbf{p}_1 - \theta\Vert_2^2 - \Vert \mathbf{p}_2 - \theta\Vert_2^2 - \Vert z - \theta \Vert^2_2.\] 

When the collective distributes mass uniformly, i.e., $h^*=(1/3,1/3,1/3)$, the centroid $\theta^*=\frac{1}{3}\sum_{i=1}^3 \mathbf{p}_i$ is a performatively stable point.
There is no incentive for either party to change their strategy since they are both best-responding to the current state.
However, a look-ahead collective would prefer $h^\sharp=(1/2,1/2,0)$, as they are aware that they could collectively shift the model $\theta$ away from this stable state further down in the triangle. This leads to a look-ahead optimal point $\theta^\sharp=\frac{1}{2}(\mathbf{p}_1+\mathbf{p}_2)$, which deviates from $\theta^*$.

In this example, \Cref{ass: linearity-of-parameter} is satisfied. Moreover, $\Phi^2>0$ since
$\Phi = -2r^2$ where $r:=\Vert\mathbf{p}_3\Vert_2$. The benefit of coordination $\mathrm{B} = U(h^\sharp) - U(h^*) = \frac{3}{4}r^2 $ is strictly positive as long as the anchor points are appropriately spaced. One can check that $U$ is $2r^2$-strongly concave, and \Cref{thm: price-of-selfish-full} can be verified since $\mathrm{B} \leq \frac{\Phi^2}{2\gamma} = r^2$ which is tight up to a factor 1/4 coming from the slack in the strong concavity assumption.

\newpage
\section{Simulations}
\label{app:SimAlign}

\subsection{Feature importance}

\begin{figure*}[h!]
  \centering
  \includegraphics[width=0.5\textwidth]{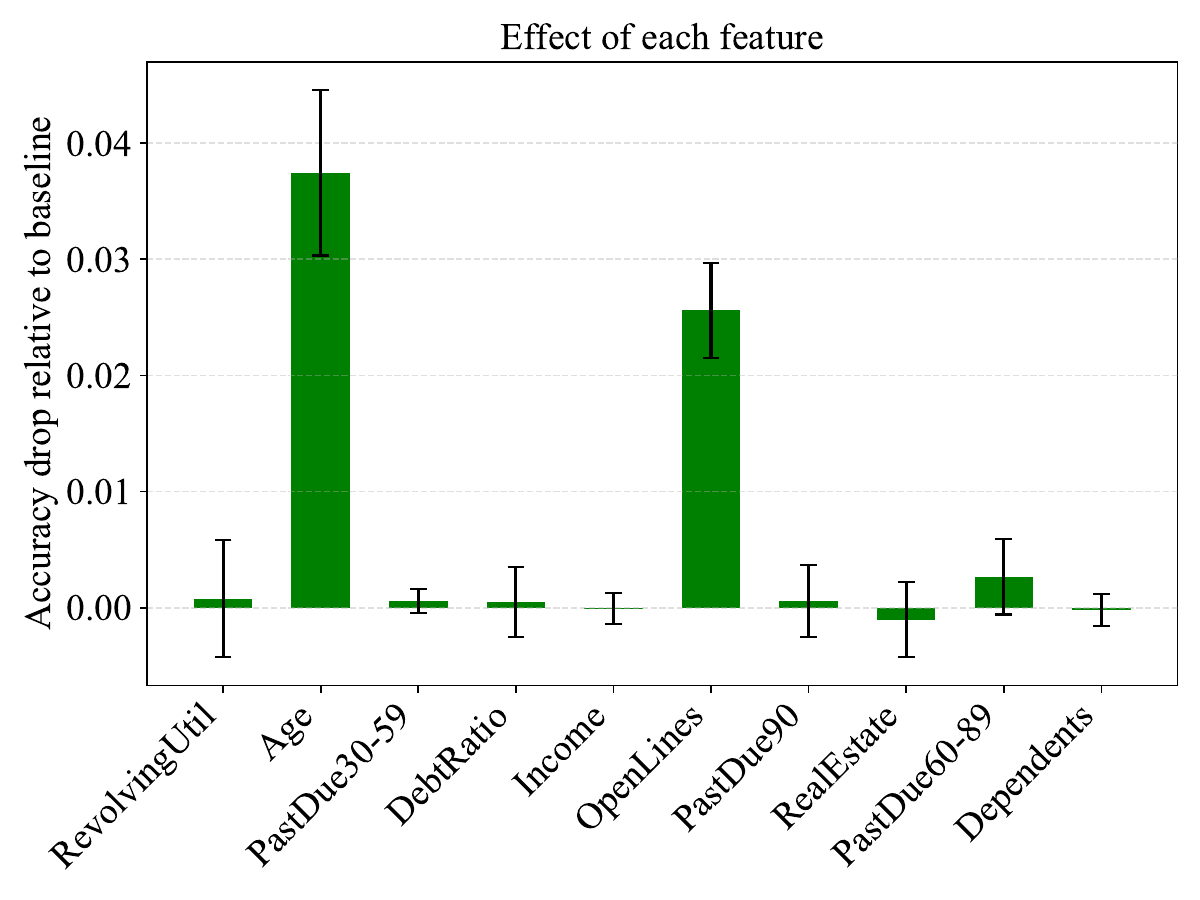}
    \caption{Accuracy drop against modifying individual features.}\label{fig:misreport}
\end{figure*}

In Figure \ref{fig:misreport} we show the importance of different features in the credit-scoring simulator. We simulate modifications to feature $i$ by replacing $z_i$ with a value ${z}_i'$, which we sample independently from a standard normal distribution. Subsequently we train a logistic regression classifier on the modified data. The values of the bars indicate the \emph{drop} of test accuracy compared to the baseline classifier trained without misreporting. The error bars indicate one standard deviations over 10 different train-test splits.

\subsection{Binary prediction setup}
\label{app:binary}

Consider the case in which the collective wants to implement a fixed strategy, which modifies their strategic features ``age'' and ``number of dependents'' as
\begin{equation*}
 h_\eta((x_i,y_i)) = (x_{i, S} + \eta\cdot \mu_{0,S} ,y_i), \label{app-eq: fixed-strategy}
\end{equation*}
where $\mu_{0,S}$ is the mean of the strategic features when the label equals $0$. Since the features are all centered across the whole dataset, this transformation can be interpreted geometrically as translating the samples with label $1$ along the direction of $\mu_0$, moving them toward the center of the distribution with label $0$ in feature space. 

We run simulations with three strategies $h^\sharp_{\alpha=0.3}$, $h^\sharp_{\alpha=0.5}$, and $h^\sharp_{\alpha=0.8}$, where each strategy optimally chooses $\eta$ for a given collective size $\alpha \in \{0.3, 0.5, 0.8\}$. The target model $\theta_{\mathrm{target}}$ we used in the simulation was fixed to be the model produced by the strategy with parameters $(\eta=0.5, \alpha=0.3)$.
For each $\alpha \in \{0.3, 0.5, 0.8\}$, we can solve for an optimal $\eta_\alpha$ such that the resulting model coincides with $\theta_{\mathrm{target}}$. In other words, for all three collective sizes, there exists a manipulation strength $\eta_\alpha$ that enables the collective to attain maximum utility (equal to $0$). This ensures that comparisons across different values of $\alpha$ are well-defined.

\end{document}

%% file: header.tex
\usepackage{amsmath,amsthm,amsfonts,amssymb,color,hyperref,anysize,enumitem,graphicx,epstopdf,algorithm,cleveref,multirow}

\usepackage{subcaption}

\newcommand{\cD}{\mathcal{D}}
\newcommand{\cZ}{\mathcal{Z}}
\newcommand{\cX}{\mathcal{X}}
\newcommand{\cY}{\mathcal{Y}}

\newcommand{\expectsub}[2]{\mathbb{E}_{#1}\left[ #2 \right]}

\DeclareMathOperator*{\argmin}{arg\,min}
\DeclareMathOperator*{\argmax}{arg\,max}

\usepackage{framed}

\newtheorem{theorem}{Theorem}
\newtheorem{lemma}[theorem]{Lemma}
\newtheorem{corollary}{Corollary}

\newtheorem{example}{Example}

\newtheorem{prop}[theorem]{Proposition}
\newtheorem{definition}{Definition}

\newtheorem{assumption}{Assumption}

\newtheorem{theoremalt}{Theorem}[theorem]

\newenvironment{theoremp}[1]{
  \renewcommand\thetheoremalt{#1}
  \theoremalt
}{\endtheoremalt}

\newcommand{\calA}{\mathcal{A}}

\newcommand{\calW}{\mathcal{W}}

\newcommand{\bbH}{\mathbf{H}}

\newcommand{\inner}[2]{\left\langle #1 , #2\right\rangle}
\newcommand{\innerH}[2]{\left\langle #1 , #2\right\rangle_{(\bbH^\star)^{-1}}}
\newcommand{\norm}[1]{\left\Vert #1 \right\Vert_2}

\definecolor{DarkGreen}{rgb}{0.075,0.375,0.075}
\definecolor{DarkRed}{rgb}{0.5,0.1,0.1}
\definecolor{DarkBlue}{rgb}{0.1,0.1,0.5}
\definecolor{Gray}{rgb}{0.2,0.2,0.2}



%% file: refs.bib
@article{zrnic2021leads,
  title={Who leads and who follows in strategic classification?},
  author={Zrnic, Tijana and Mazumdar, Eric and Sastry, Shankar and Jordan, Michael},
  journal={Advances in Neural Information Processing Systems},
  volume={34},
  pages={15257--15269},
  year={2021}
}

@inproceedings{perdomo20performativeprediction,
  title={Performative prediction},
  author={Perdomo, Juan and Zrnic, Tijana and Mendler-D{\"u}nner, Celestine and Hardt, Moritz},
  booktitle={International Conference on Machine Learning},
  pages={7599--7609},
  year={2020}
}

@inproceedings{dong2018strategic,
  title={Strategic classification from revealed preferences},
  author={Dong, Jinshuo and Roth, Aaron and Schutzman, Zachary and Waggoner, Bo and Wu, Zhiwei Steven},
  booktitle={Proceedings of the 2018 ACM Conference on Economics and Computation},
  pages={55--70},
  year={2018}
}

@article{podimata2025incentive,
  title={Incentive-aware machine learning; robustness, fairness, improvement \& causality},
  author={Podimata, Chara},
  journal={arXiv preprint arXiv:2505.05211},
  year={2025}
}

@inproceedings{hardt16strategic,
author = {Hardt, Moritz and Megiddo, Nimrod and Papadimitriou, Christos and Wootters, Mary},
title = {Strategic Classification},
year = {2016},
booktitle = {ACM Conference on Innovations in Theoretical Computer Science},
pages = {111–122},
numpages = {12},
}

@article{chen2020learning,
  title={Learning strategy-aware linear classifiers},
  author={Chen, Yiling and Liu, Yang and Podimata, Chara},
  journal={Advances in Neural Information Processing Systems},
  volume={33},
  pages={15265--15276},
  year={2020}
}

@Article{nagel95,
  author={Nagel, Rosemarie},
  title={{Unraveling in Guessing Games: An Experimental Study}},
  journal={American Economic Review},
  year=1995,
  volume={85},
  number={5},
  pages={1313-1326}
}

@inproceedings{bruckner2011stackelberg,
  title={Stackelberg games for adversarial prediction problems},
  author={Br{\"u}ckner, Michael and Scheffer, Tobias},
  booktitle={ACM SIGKDD},
  pages={547--555},
  year={2011}
}

@inproceedings{hardt2023algorithmic,
  title={Algorithmic Collective Action in Machine Learning},
  author={Hardt, Moritz and Mazumdar, Eric and Mendler-D{\"u}nner, Celestine and Zrnic, Tijana},
  booktitle={International Conference on Machine Learning},
  articleno = {510},
  year={2023}
}

@inproceedings{mendler20stochasticPP,
        author = {Mendler-D\"{u}nner, Celestine and Perdomo, Juan and Zrnic, Tijana and Hardt, Moritz},
        booktitle = {Advances in Neural Information Processing Systems},
        pages = {4929--4939},
        title = {Stochastic Optimization for Performative Prediction},
        volume = {33},
        year = {2020}
}

@inproceedings{jag21alt,
        author = {Meena Jagadeesan and Celestine Mendler{-}D{\"{u}}nner and Moritz Hardt},
        booktitle = {International Conference on Machine Learning},
        title = {Alternative Microfoundations for Strategic Classification},
        year = {2021},
        pages = 	 {4687--4697},
        volume = 	 {139}
}

@article{drusvyatskiy23stochastic,
author = {Drusvyatskiy, Dmitriy and Xiao, Lin},
title = {Stochastic Optimization with Decision-Dependent Distributions},
journal = {Mathematics of Operations Research},
volume = {48},
number = {2},
pages = {954-998},
year = {2023},
}

@InProceedings{brown22stateful,
  title = 	 { Performative Prediction in a Stateful World },
  author =       {Brown, Gavin and Hod, Shlomi and Kalemaj, Iden},
  booktitle = 	 {International Conference on Artificial Intelligence and Statistics},
  pages = 	 {6045--6061},
  year = 	 {2022},
  volume = 	 {151}

}

@inproceedings{hardt2022power,
  title={Performative power},
  author={Hardt, Moritz and Jagadeesan, Meena and Mendler-D{\"u}nner, Celestine},
  booktitle={Advances in Neural Information Processing Systems},
  volume={35},
  pages={22969--22981},
  year={2022}
}

@article{narang23multiplayer,
  author  = {Adhyyan Narang and Evan Faulkner and Dmitriy Drusvyatskiy and Maryam Fazel and Lillian J. Ratliff},
  title   = {Multiplayer Performative Prediction: Learning in Decision-Dependent Games},
  journal = {Journal of Machine Learning Research},
  year    = {2023},
  volume  = {24},
  number  = {202},
  pages   = {1--56}
}

@article{baumann2024algorithmic,
  title={Algorithmic collective action in recommender systems: promoting songs by reordering playlists},
  author={Baumann, Joachim and Mendler-D{\"u}nner, Celestine},
  journal={Advances in Neural Information Processing Systems},
  volume={37},
  pages={119123--119149},
  year={2024}
}

@inproceedings{sigg2024decline,
  title={Decline now: A combinatorial model for algorithmic collective action},
  author={Sigg, Dorothee and Hardt, Moritz and Mendler-D{\"u}nner, Celestine},
  booktitle={Proceedings of the 2025 CHI Conference on Human Factors in Computing Systems},
  pages={1--17},
  year={2025}
}

@InProceedings{gauthier2025statisticalcollusioncollectiveslearning,
  title = 	 {Statistical Collusion by Collectives on Learning Platforms},
  author =       {Gauthier, Etienne and Bach, Francis and Jordan, Michael I.},
  booktitle = 	 {Proceedings of the 42nd International Conference on Machine Learning},
  pages = 	 {18897--18919},
  year = 	 {2025},
  volume = 	 {267}
}

@inproceedings{ben2024role,
  title={The Role of Learning Algorithms in Collective Action},
  author={Ben-Dov, Omri and Fawkes, Jake and Samadi, Samira and Sanyal, Amartya},
  booktitle={International Conference on Machine Learning},
  pages={3443--3461},
  year={2024}
}

@inproceedings{balduzzi2018mechanics,
  title={The mechanics of n-player differentiable games},
  author={Balduzzi, David and Racaniere, Sebastien and Martens, James and Foerster, Jakob and Tuyls, Karl and Graepel, Thore},
  booktitle={International Conference on Machine Learning},
  pages={354--363},
  year={2018}
}

@inproceedings{bechavod2021information,
  title={Information discrepancy in strategic learning},
  author={Bechavod, Yahav and Podimata, Chara and Wu, Steven and Ziani, Juba},
  booktitle={International Conference on Machine Learning},
  pages={1691--1715},
  year={2022}
}

@InProceedings{ghalme2021strategic,
  title = 	 {Strategic Classification in the Dark},
  author =       {Ghalme, Ganesh and Nair, Vineet and Eilat, Itay and Talgam-Cohen, Inbal and Rosenfeld, Nir},
  booktitle = 	 {Proceedings of the 38th International Conference on Machine Learning},
  pages = 	 {3672--3681},
  year = 	 {2021},
  volume = 	 {139}
}

@article{hardt25sts,
author = {Moritz Hardt and Celestine Mendler-D{\"u}nner},
title = {{Performative Prediction: Past and Future}},
volume = {40},
journal = {Statistical Science},
number = {3},
publisher = {Institute of Mathematical Statistics},
pages = {417 -- 436},
year = {2025}
}

@article{tassinari20solidarity,
author = {Arianna Tassinari and Vincenzo Maccarrone},
title ={Riders on the Storm: Workplace Solidarity among Gig Economy Couriers in Italy and the UK},
journal = {Work, Employment and Society},
volume = {34},
number = {1},
pages = {35-54},
year = {2020}
}

@article{chen18didi,
author = {Julie Yujie Chen},
title ={Thrown under the bus and outrunning it! The logic of Didi and taxi drivers’ labour and activism in the on-demand economy},
journal = {New Media \& Society},
volume = {20},
number = {8},
pages = {2691-2711},
year = {2018}
}

@article{kneeland15,
author = {Kneeland, Terri},
title = {Identifying Higher-Order Rationality},
journal = {Econometrica},
volume = {83},
number = {5},
pages = {2065-2079},
year = {2015}
}

@inproceedings{vincent21dataleverage,
author = {Vincent, Nicholas and Li, Hanlin and Tilly, Nicole and Chancellor, Stevie and Hecht, Brent},
title = {Data Leverage: A Framework for Empowering the Public in its Relationship with Technology Companies},
year = {2021},
booktitle = {Proceedings of the 2021 ACM Conference on Fairness, Accountability, and Transparency},
pages = {215–227},
numpages = {13}
}

@misc{drivers-united,
    key = {Rideshare Drivers United},
    note = {\url{https://www.drivers-united.org/}}
}

@InProceedings{levanon22ageneralSC,
  title = 	 {Generalized Strategic Classification and the Case of Aligned Incentives},
  author =       {Levanon, Sagi and Rosenfeld, Nir},
  booktitle = 	 {Proceedings of the 39th International Conference on Machine Learning},
  pages = 	 {12593--12618},
  year = 	 {2022},
  volume = 	 {162},
  publisher =    {PMLR},

}

@inproceedings{kulynych20POT,
author = {Kulynych, Bogdan and Overdorf, Rebekah and Troncoso, Carmela and G\"{u}rses, Seda},
title = {POTs: protective optimization technologies},
year = {2020},
isbn = {9781450369367},
publisher = {Association for Computing Machinery},
booktitle = {Proceedings of the 2020 Conference on Fairness, Accountability, and Transparency},
pages = {177–188},
numpages = {12}
}

@inproceedings{
nestaas2025adversarial,
title={Adversarial Search Engine Optimization for Large Language Models},
author={Fredrik Nestaas and Edoardo Debenedetti and Florian Tram{\`e}r},
booktitle={The Thirteenth International Conference on Learning Representations},
year={2025}
}
